\documentclass[10pt,twocolumn,letterpaper]{article}

\usepackage{cvpr}
\usepackage{times}
\usepackage{graphicx}
\usepackage{amsmath}
\usepackage{amssymb}
\usepackage{amsmath}
\usepackage{tabularx}
\usepackage{url}
\usepackage{color}
\usepackage{xspace}
\usepackage{balance}
\usepackage{cleveref}

\crefformat{footnote}{#2\footnotemark[#1]#3}

\def\@onedot{\ifx\@let@token.\else.\null\fi\xspace}

\makeatother

\newcommand{\dataset}{\boldsymbol{X}}
\newcommand{\classlabels}{\boldsymbol{C}}
\newcommand{\image}{\boldsymbol{I}}

\newcommand{\vecx}{\boldsymbol{x}}
\newcommand{\vecw}{\boldsymbol{w}}

\usepackage[pagebackref=false,breaklinks=true,letterpaper=true,colorlinks,citecolor=blue,bookmarks=false]{hyperref}

\cvprfinalcopy 

\def\cvprPaperID{****} 

\ifcvprfinal\fi
\begin{document}

\title{Subset Feature Learning for Fine-Grained Category Classification}

\author
  {
  {\it ZongYuan Ge{\tiny ~}$^{\dagger\ddagger}$, Christopher McCool{\tiny ~}$^{\ddagger}$, Conrad Sanderson{\tiny ~}$^{\ast\diamond}$, Peter Corke{\tiny ~}$^{\dagger\ddagger}$}\\
  ~\\
  $^\dagger$     Australian Centre for Robotic Vision, Brisbane, Australia\\
  $^\ddagger$  Queensland University of Technology (QUT), Brisbane, Australia\\
  $^\ast$ University of Queensland, Brisbane, Australia\\
  $^\diamond$ NICTA, Australia\\
  }


\maketitle
\thispagestyle{empty}

\begin{abstract}
\vspace{-1ex}

Fine-grained categorisation has been a challenging problem due to
small inter-class variation, large intra-class variation and low number of training images. 
We propose a learning system which first clusters visually similar classes
and then learns deep convolutional neural network features specific to each subset.
Experiments on the popular fine-grained Caltech-UCSD bird dataset show that the proposed 
method outperforms recent fine-grained categorisation methods under 
the most difficult setting: no bounding boxes are presented at test time.
It achieves a mean accuracy of $77.5\%$, compared to the previous best performance of $73.2\%$.
We also show that progressive transfer learning allows us to first learn 
domain-generic features (for bird classification) which can then be 
adapted to specific set of bird classes, yielding improvements in accuracy.

\end{abstract}

\vspace{-2ex}
\section{Introduction}
\label{sec:introduction}



Deep convolutional neural networks (CNNs) have been successful in various computer vision tasks. 
Deep CNNs have achieved impressive in both general~\cite{krizhevsky2012imagenet,razavian2014cnn,donahue2013decaf}
and fine-grained image classification~\cite{zhang2014part,ge2015modelling}. 
Recently, deep CNN approaches have been shown to surpass human performance for 
the task of recognising 1000 classes from the ImageNet dataset~\cite{he2015delving}. 
Although deep CNNs can serve as an end-to-end classifier, they have been used by 
many researchers as a feature extractor for various recognition problem 
including segmentation~\cite{hariharan2014simultaneous} and detection~\cite{girshick2013rich}. 

Recently, the task of fine-grained image categorisation has received considerable attention,
in particular the task of fine-grained bird classification~\cite{zhang2014part,berg2013poof,chai2013symbiotic,farrell2011birdlets,gavves2013fine}. 
Fine-grained image classification is a challenging computer vision problem due 
to subtle differences in the overall appearance between various classes (low inter-class variation)
and large pose and appearance variations in the same class (large intra-class variation).

Much of the work for fine-grained image classification has dealt with the issue 
of detecting and modelling local parts.
Several researchers have examined methods to find local parts and extract 
normalised features in order to overcome the issues of pose and view-point 
variation~\cite{branson2011strong,chai2013symbiotic,liu2012dog,zhang2013deformable,donahue2013decaf}. 
Aside from the issue of pose and view-point changes, a major challenge for any 
fine-grained classification approach is how to distinguish between classes that have high visual correlations~\cite{berg2013poof}. 
Some state-of-the-art pose normalised methods still have considerable difficulty in categorising some visually similar fine-grained classes~\cite{zhang2014part,branson2014bird}.

To date, there has been limited work which investigates in detail how best to learn deep CNN features for the fine-grained classification problem. 
Most of the methods used off-the-shelf convolutional neural networks (CNNs) features trained from ImageNet or fine-tuned the pre-trained ImageNet model
on the target dataset, then using one fully-connected layer as a feature descriptor~\cite{jia2014caffe,razavian2014cnn}. 

This paper examines in detail how to best learn deep CNN features for fine-grained image classification.
In doing so, we propose a novel \textit{subset} learning system which first splits the classes into visually similar subsets and then learns domain-specific features for each subset.
We also comprehensively investigate progressive transfer learning and highlight that first learning domain-generic features (for bird classification) 
using a large dataset and then adapting this to the specific task (target bird dataset) yields considerable performance improvements.
 

\begin{figure}[!tb]
  \centering
  \includegraphics[width=1\linewidth]{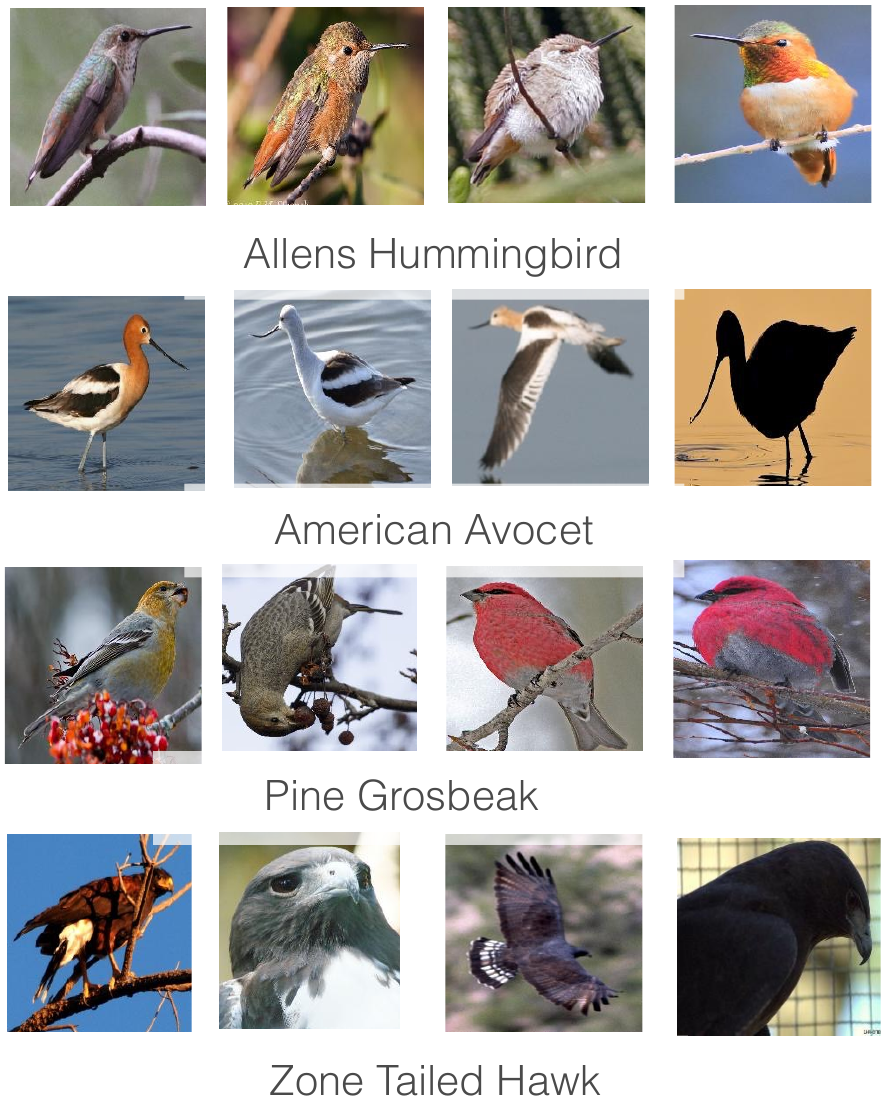}
  \caption
    {
    Birdsnap is a very challenging fine-grained bird dataset with sexual as well as age dimorphisms.
    There are considerable appearance differences between males and females,
    as well as between young and mature birds.
    Each row shows images from the same species.
    For each bird species there are large intra-class variations: pose variation, background variation and appearance variation.
    }
  \label{fig:birdsnap}
\end{figure}

\section{Related Work}
\label{sec:related_work}


\subsection{Convolutional Neural Networks}

Krizhevsky et al.~\cite{krizhevsky2012imagenet} recently achieved impressive performance on the ImageNet recognition task using CNNs,
which were initially proposed by LeCun et al.~\cite{LeCun89} for hand writing digit recognition.
Since then CNNs have received considerable attention~\cite{razavian2014cnn,girshick2013rich}.
The network structure of Krizhevsky et al.~\cite{krizhevsky2012imagenet} remains a popular structure and consists of five convolutional layers ($conv1$ to $conv5$) with two fully-connected layers ($fc6$ and $fc7$) followed by a softmax layer to predict the class label.
The network is capable of generating useful feature representations by learning low level features in early convolutional layers and 
accumulating them to high level semantic features in the latter convolutional layers~\cite{zeiler2013visualizing}. 


\subsection{Features for Fine-grained Classification}

Several approaches have been designed to learn feature representations for fine-grained image classification. 
Berg et al.~\cite{berg2013poof} generated millions of keypoint pairs to learn a set of highly discriminative features.
Zhang et al.~\cite{zhang2013deformable} learned pose normalised features by using the deformable part descriptors model (DPM)~\cite{felzenszwalb2008discriminatively}
on local parts which were extracted using a pre-trained deep CNN. 
Chen et al.~\cite{chen2015selective} proposed a framework to select the most confident local descriptors for nonlinear function learning
using a linear approximation in an embedded higher dimensional space.

The above feature learning schemes are implicitly part-based methods.
This means they require the ground truth locations of each part which limits their usefulness in terms of fully automatic deployment.

\section{Proposed Method}
\label{sec:proposed_method}



Our proposed feature learning method consists of two main parts.
First, we perform progressive transfer learning to learn a domain-generic convolutional feature extractor (termed $\phi_{GCNN}$) from a large-scale dataset of the same domain as the target dataset.
Second, we perform subset-specific feature learning from pre-clustered subsets which contain visually similar fine-grained class images.
The discriminative convolutional features learned from the subset learning system is termed $DFCNN$, and the related feature extractor is referred as $\phi_{DFCNN}$. 

For image $\image_{i}$, we apply the $\phi_{GCNN}(\image_{i})$ and $\phi_{DFCNN}(\image_{i})$ and combine them to obtain our feature vector to describe the image.
For training the classifier, we employ a one-versus-all linear SVM using the final feature representation. 

\subsection{Progressive Transfer Learning}

It is desirable to have as much as data possible in order to avoid overfitting while training a CNN. 
A typical CNN has millions of parameters which makes it difficult to train when data is limited. 
Typically fine-grained image datasets are relatively small compared to the ImageNet dataset.
To circumvent problems with small datasets, a process known as transfer learning~\cite{yosinski2014transferable} can be applied.
Transfer learning has usually been applied by fine-tuning a general network, such as the network of Krizhevsky et al.~\cite{krizhevsky2012imagenet}, to a specific task such as bird classification~\cite{zhang2014part}.
Recent work by Yosinski et al.~\cite{yosinski2014transferable} found that better accuracy can be achieved if transfer learning is performed using datasets representing the same or related domains.

Inspired by the findings of Yosinski et al.~\cite{yosinski2014transferable}, we propose an alternative approach where a generic CNN is progressively adapted to the task at hand.
First, a large dataset, which is related to the same domain as the final task, is used to perform transfer learning.
This yields a domain-generic feature representation.
Second, a smaller dataset which represents the final task at hand is used to adapt the domain-generic features to yield task-specific features.
Our experimental results show that progressive transfer learning yields feature representation which lead to consistently improved performance.
Furthermore, we will show that the domain-generic features can also be used effectively for the task at hand.



\subsection{Subset Specific Feature Learning}

Recent parts-based fine-grained methods show relatively good performance on the Caltech-UCSD bird dataset~\cite{wah2011caltech}.
The methods are good at recognising birds species with distinguishable features with moderate pose variation.
However, many mis-classifications occur for birds species that have similar visual appearance. 

\begin{figure}[!tb]
  \centering
  \includegraphics[width=1\linewidth]{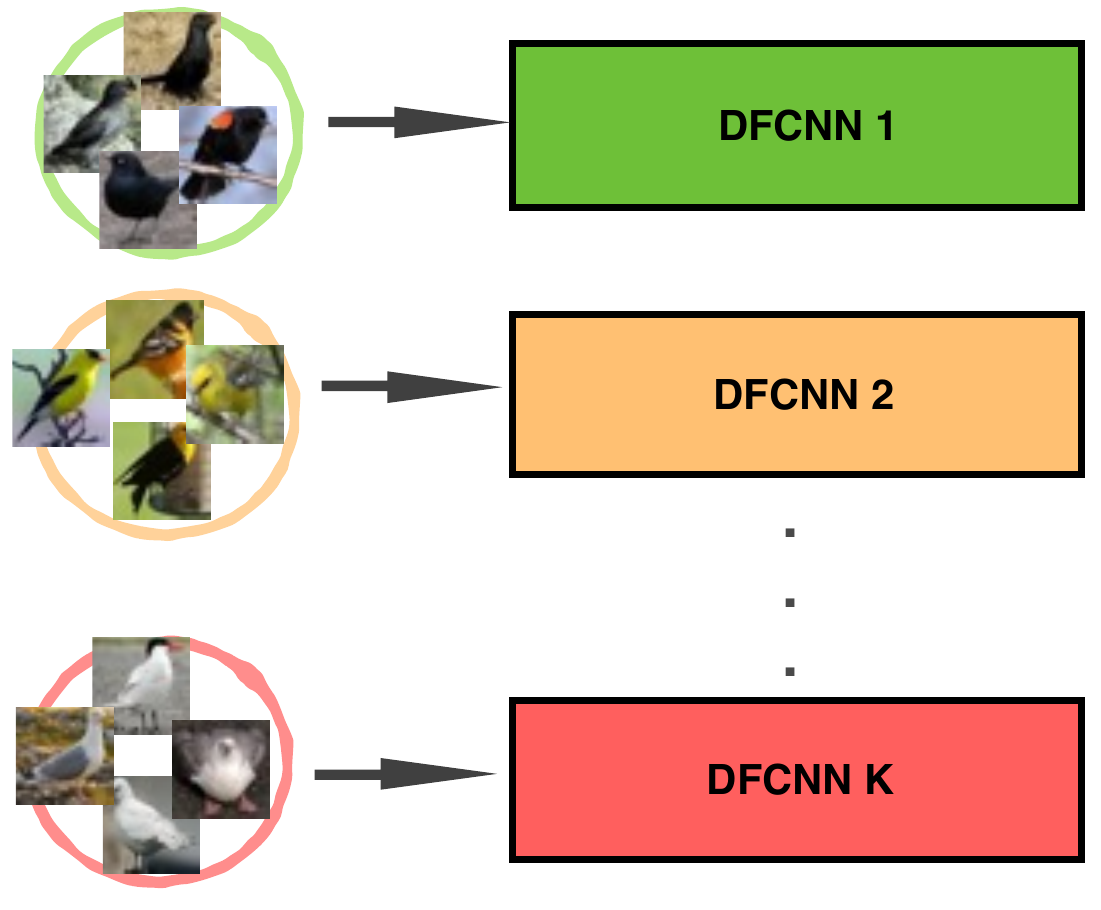}
  \caption
    {
    Pre-clustered visually similar images are fed into $DFCNN_{1...K}$ with backpropogation training to learn discriminative features for each subset. 
    }
  \label{fig:subset_fea}
\end{figure}

To address this issue, we propose to pre-cluster visually similar species into subsets and use subset-specific CNNs.
Instead of relying on one CNN to handle all possible cases,
each CNN focuses on the differences within each subset.
In effect, the overall classifier has more parameters, as all CNNs have the same network architecture.
Due to the practical issues such as training time and memory requirements,
using separate CNNs dedicated to specific tasks is more practical than having one very large CNN.
An overview of this subset learning scheme is shown in Fig.~\ref{fig:subset_fea}.

The above subset feature learning process is initially performed on a large yet related dataset.
In particular, we use the large Birdsnap dataset~\cite{berg2014birdsnap} instead of the target Caltech-UCSD dataset~\cite{wah2011caltech}.
We expect that our learned features are both generalised and discriminative compared to features learned directly
on the same size or smaller size target dataset under the same domain. 

\subsubsection{Pre-clustering}

To generate subsets in terms of visually similar images,
image representations should focus on colour and texture while being robust to pose and background variations.
We investigate three types of features as image representers.
Features are obtained from either the 5-th layer $conv5$ or the 6-th layer ($fc6$) of the CNN.
These were selected due to their recent use by other researchers to perform object recognition and clustering~\cite{donahue2013decaf}.
We also apply linear discriminant analysis (LDA)~\cite{rao1948utilization} to $fc6$ features to reduce their dimensionality.
This is done to ameliorate the well known issues of clustering high dimensional data~\cite{aggarwal2005k}. 
The subsets are then obtained via $k$-means clustering.



\begin{figure}[!tb]
  \centering
  
  \begin{minipage}{1\columnwidth}
    \begin{minipage}{0.05\columnwidth}
    {\bf (a)}
    \end{minipage} 
    \begin{minipage}{0.92\columnwidth}
      \includegraphics[width=1\linewidth]{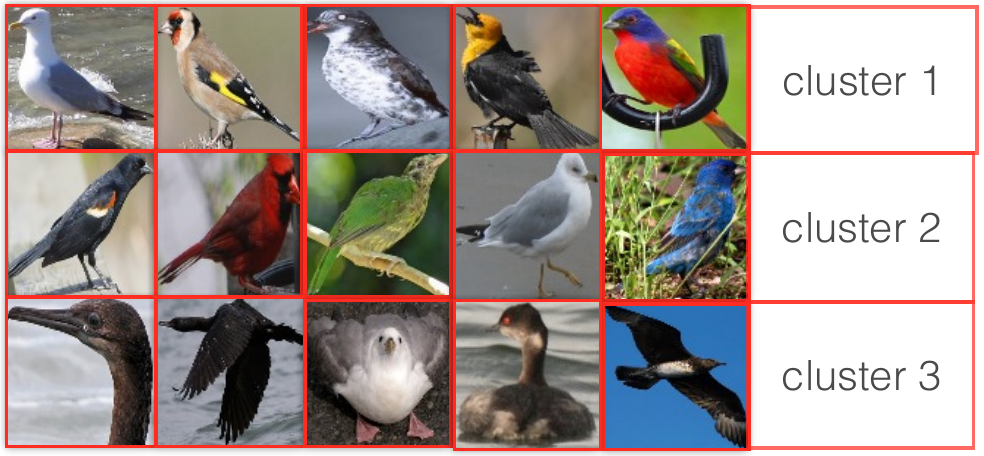}
    \end{minipage}
  \end{minipage}
  
  \vspace{1ex}
  \hrule
  \vspace{1ex}
  
  \begin{minipage}{1\columnwidth}
    \begin{minipage}{0.05\columnwidth}
    {\bf (b)}
    \end{minipage} 
    \begin{minipage}{0.92\columnwidth}
      \includegraphics[width=1\linewidth]{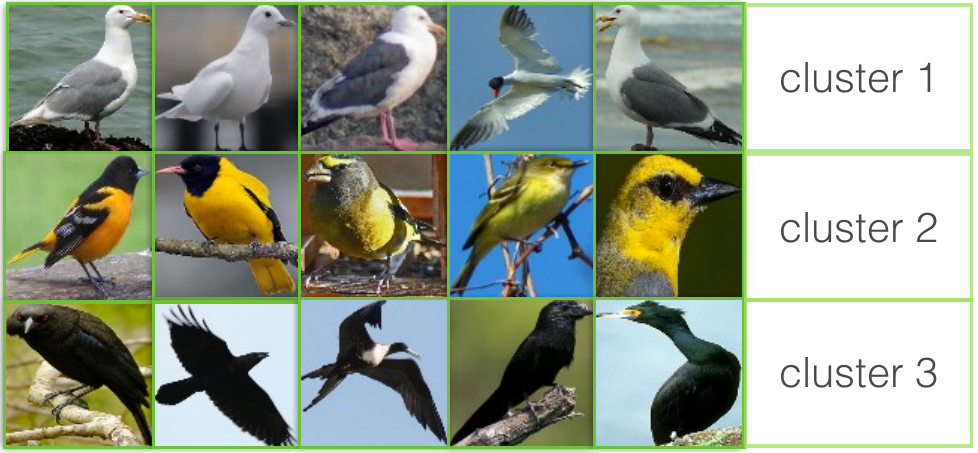}
    \end{minipage}
  \end{minipage}

  \vspace{1ex}
  \hrule
  \vspace{1ex}

  \begin{minipage}{1\columnwidth}
    \begin{minipage}{0.05\columnwidth}
    {\bf (c)}
    \end{minipage} 
    \begin{minipage}{0.92\columnwidth}
      \includegraphics[width=1\linewidth]{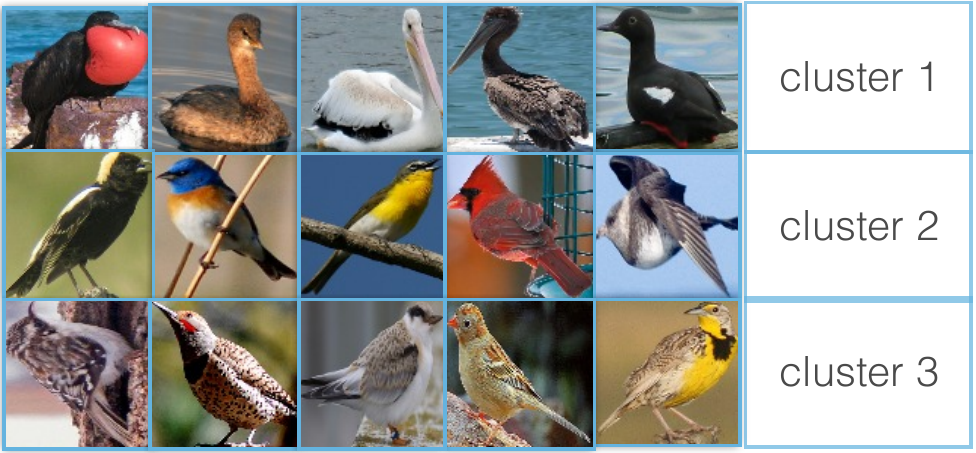}
    \end{minipage}
  \end{minipage}
  
  \vspace{1ex}
  
  \caption
    {
    Pre-clustering results using:
    {\bf (a)}~$conv5$ layer features,
    {\bf (b)}~$fc6$ layer features,
    {\bf (c)}~$lda-fc6$ features.
    Clustering via $conv5$ yields undesirable strong correlations with pose and shape information.
    Using $fc6$ yields some improvements, but the pose bias is still visibly present.
    Using $lda-fc6$ provides further clustering improvements in terms of robustness to color and pose variations.
    }
  \label{fig:cluster}
\end{figure}

Examples of clustering results using the three feature types are shown in Fig.~\ref{fig:cluster}.
The fully connected layer based feature $fc6$ fits our criteria better than 
clustering using the the convolutional feature $conv5$ that tends to learn shape and pose information, which is undesirable.
This particular property can be seen in clusters 1 and 2 in Fig.~\ref{fig:cluster}(a) which represent right and left pose of birds images while the rest are grouped into cluster 3.
We conjecture that this is due to the convolutional based features containing a high degree of spatial information.
Using $fc6$ yields some improvements, but the pose bias is still visibly present.
Using $lda-fc6$ features provides further clustering improvements in terms of robustness to colour and pose variations.

\subsubsection{Subset Feature Learning}


A separate CNN is learned for each of the $K$ pre-clustered subsets.
The aim is to learn features for each subset that will allow us to more easily differentiate visually similar species.
As such, for each subset, we apply transfer learning to the CNN of Krizhevsky et al.~\cite{krizhevsky2012imagenet} (whose structure was described in Section~\ref{sec:related_work}).
To train the $k$-th subset ($Subset_{k}$) we use the $N_{k}$ images assigned to this subset $\dataset_{k}=\left[\vecx_{1}, \dots , \vecx_{N_{k}} \right]$, with their corresponding class labels $\classlabels_{k}=\left[c_{1}, \dots , c_{N_{k}} \right]$.
The number of outputs in the associated last fully connected layer $fc8$ is set to the number of classes in each subset.
Transfer learning is then applied separately to each network using backpropogation and stochastic gradient descent (SGD).
We then take $fc6$ to be the learned subset feature $\phi_{DFCNN_{k}}$ for the $k$-th subset.


\subsection{Fine-grained Classification}

\begin{figure}[!tb]
  \centering
  \includegraphics[width=1\columnwidth]{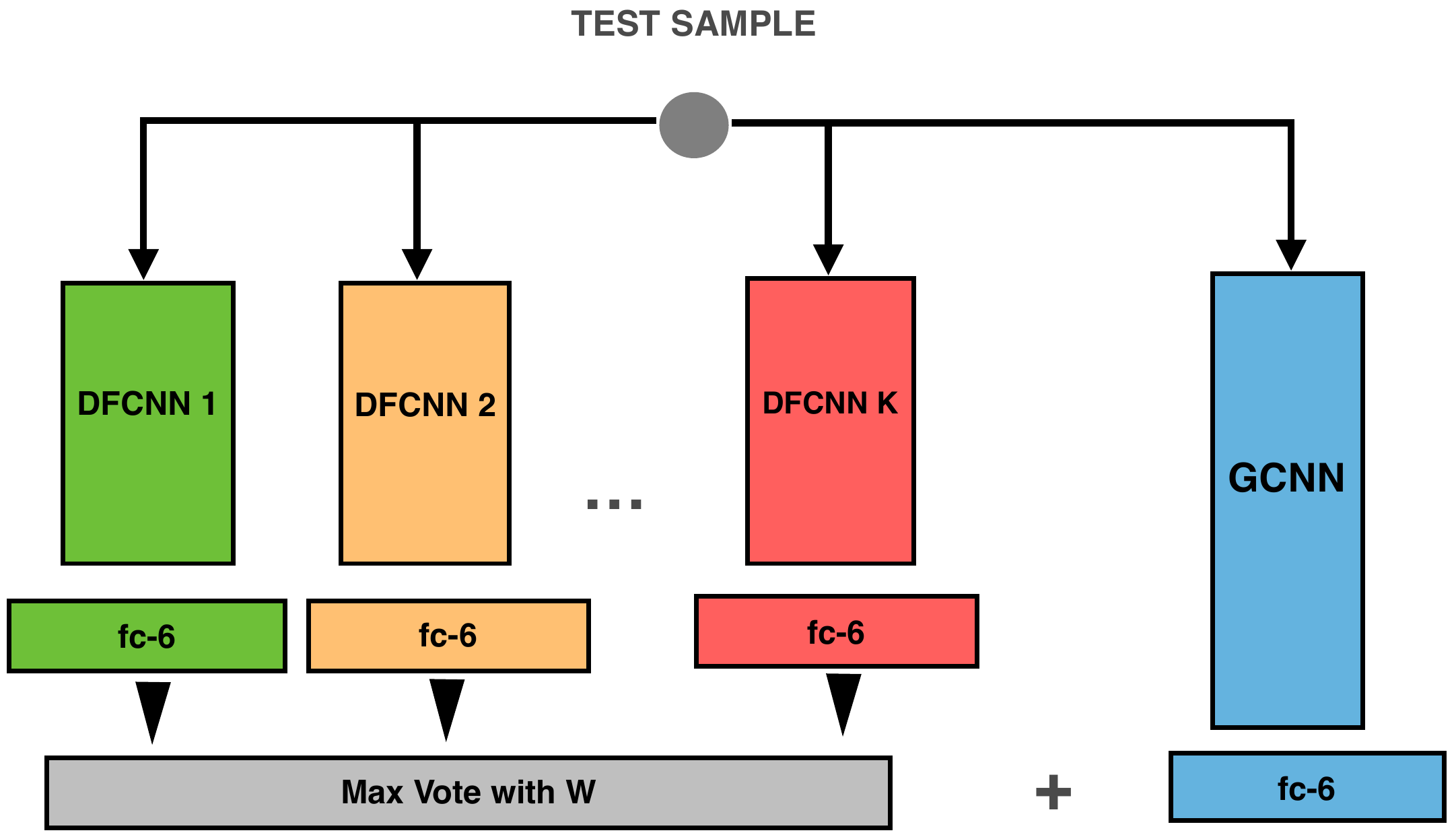}
  \caption
    {Feature representation of the test image is the concatenated features from both DFCNN with weighting factors and GCNN.
    }
  \label{fig:fea}
\end{figure}

To predict test labels for an image $\image_{t}$, our classification pipeline combines the $\phi_{GCNN}(\image_{t})$ feature with the $K$ subset features $\phi_{DFCNN_{1...K}}(\image_{t})$.
A max voting rule is used to retain only the most relevant subset-specific feature.
The other $K-1$ features are set to $\mathbf{0}$.
See Fig.~\ref{fig:fea} for a conceptual representation.
To balance weights for the domain-generic and subset-specific features, both $GCNN$ and $DFCNN$ features are then $l2$ normalised before combining them into a single feature vector.
Using this feature vector, we train a one-versus-all linear SVM in order to make predictions.


\vspace{-0.5ex}
\subsubsection{Max Voting DFCNN}

The final feature representation for image $\image$ is the concatenation of generalised features obtained from $\phi_{GCNN}(\image)$ and the $K$ subsets $\phi_{DFCNN_{1...K}}(\image)$.
However, sometimes an image is more relevant to one subset features than others. 
For example to extract features for a White Gull image, it is more reasonable to use $DFCNN$ features from the subset which has many relevant white birds.

To emphasise the most relevant $DFCNN$, we first learn a \textbf{subset selector} to select the most relevant subset (rank 1) to the image.
Max voting is then used to retain the feature from the most relevant subset and the remaining $k-1$ subset features are set to $0$.
One way to interpret the max voting is to use the \textbf{subset selector} to learn a binary vector $\vecw$, where $\sum_{i=1}^K \vecw_{i} = 1$. 
The final subset feature representation is then $DFCNN = [w_{1}\phi_{DFCNN_{1}}(x_{i}), \dots , w_{k}\phi_{DFCNN_{K}}(x_{i})]$.  
We explore two ways to learn the \textbf{subset selector}.

The simplest way of learning the \textbf{subset selector} is to use the centroids from the pre-clustering; we refer to this as $Cen_{1...K}$.
This provides a simple classifier trained in an unsupervised manner, however, given the importance of this stage we explore the use of a discriminatively trained classifier using a CNN.

Another way to select the most relevant subset is to train a separate CNN based subset selector $SCNN$.
Using the output from the pre-clustering as the class labels, we learn a new SCNN by changing the softmax layer $fc8$ to have $K$ outputs.
The softmax layer now predicts the probability of the test image belonging to a specific subset $Subset_{k}$, max voting is then applied to this prediction to choose the most likely subset.
As with the previously trained CNNs, the weights of $SCNN$ are trained via backpropogation and SGD using the network of Krizhevsky et al.~\cite{krizhevsky2012imagenet} as the starting point.

%
%
%

\section{Experiments}
\label{sec:experiment}

In this section we present a comparative performance evaluation of our proposed method. 
We conduct experiments on the Caltech-UCSD dataset~\cite{wah2011caltech},
which is the most widely used benchmark for fine-grained classification.
We train the model using ImageNet~\cite{krizhevsky2012imagenet}
and recently released Birdsnap dataset~\cite{berg2014birdsnap}.

ImageNet consists of 1000 classes with approximately 1000 images for each class. 
In total there are approximately 1.2 million training images.

Caltech-UCSD contains 11,788 images across 200 species. 
Birdsnap contains 500 species of North American birds with 49,829 images.
Examples are shown in Fig.~\ref{fig:birdsnap}.
Birdsnap is similar in structure to Caltech-UCSD, but has several differences.
First, it contains overlapping 134 species and four times the number of images than Caltech-UCSD. 
Second, there is strong intra-variation within many species due to sexual as well as age dimorphisms.
There are considerable appearance differences between males and females, as well as between young and mature birds.

We use the implementation of LDA and $k$-means from the Bob library~\cite{bob2012}.
The open-source package Caffe~\cite{jia2014caffe} is used to train and extract CNN features. 
We use $lda-fc6$ layer features to pre-cluster subsets and $fc6$ features for classification.


\subsection{Evaluation of Transfer Learning for Domain-Generic Features}

The CNN model architecture is identical to the model used by Krizhevsky et al.~\cite{krizhevsky2012imagenet}. 
We fine-tune the CNN model by using training images from the ground truth bounding box crops of the original images.
The resultant cropped images are all resized $227 \times 227$. 
During test time, ground truth bounding box crops of the test images from Caltech-UCSD are used to make predictions. 

\newpage
We conducted 3 sets of experiments for transfer learning:

\begin{enumerate}

\item
The first experiment used all of the data from Birdsnap (500 species)
to perform large-scale progressive feature learning.

\item
In the second experiment we removed those species in Birdsnap and Caltech-UCSD that overlapped. 
This allows us to examine the potential for learning domain features that are not specific to the task at hand.

\item
In the third experiment we explored the impact that including the overlapping species has on the transfer learning process.

\end{enumerate}

We use the following acronyms.
\textbf{IN} represents using weights from the pre-trained ImageNet model.
We define \textbf{rt} as retraining the network from scratch with random initialised weights.
\textbf{ft} refers to fine-tuning the network.
For example, \textbf{IN-CUB-ft} means fine-tuning the ImageNet model weights on the Caltech-UCSD bird dataset. 
ImageNet dataset is represented as \textbf{IN},
while Caltech-UCSD is \textbf{CUB}, and Birdsnap is \textbf{BS}.

\vspace{-1ex}
\subsubsection{Transfer Learning: Experiment I}

In this experiment we used all images (500 species) from Birdsnap
to explore large-scale progressive feature learning. 
We exclude those images that exist in both Birdsnap and the Caltech-UCSD datasets. 

The first three rows of Table~\ref{table:transfer} show the accuracy when the 
CNNs are trained from scratch.
In this setting the \textbf{IN-rt} system, the pre-trained network generated by 
Krizhevsky et al.~\cite{krizhevsky2012imagenet} on ImageNet, performs the best 
with a mean accuracy of $58.0\%$.
Interestingly, the \textbf{BS-rt} system has a considerably higher mean accuracy 
of $44.8\%$ when compared to \textbf{CUB-rt} which has a mean accuracy of 
$11.4\%$.
We believe that this indicates that the Birdsnap dataset has almost enough data 
to train a deep CNN from scratch.

Transfer learning offers a way to mitigate the lack of sufficient domain data.
As such, we performed transfer learning by fine-tuning the pre-trained CNN. 
We did this using just the Caltech-UCSD (target) dataset \textbf{IN-CUB-ft} or 
the Birdsnap (domain specific) dataset \textbf{IN-BS-ft}.

Somewhat surprisingly, training on the target dataset (\textbf{IN-CUB-ft}) 
provides a lower mean accuracy of $68.3\%$ when compared to using the domain 
specific dataset (\textbf{IN-BS-ft}) which has a mean accuracy of $70.1\%$.
Performing progressive feature learning on the \textbf{IN-BS-ft} CNN leads to 
further improvements achieving a mean accuracy of $70.8\%$ 
(\textbf{IN-BS-ft-CUB-ft}).
These two results demonstrate the potential for learning domain-generic features 
(\textbf{IN-BS-ft}) as well as progressive feature learning to perform effective 
transfer learning (\textbf{IN-BS-ft-CUB-ft}) for fine-grained image 
classification.

An obvious issue that is not addressed in this first experiment is that there 
are overlapping species in Birdsnap and Caltech-UCSD.
To evaluate the impact of this we perform two more experiments.

\begin{table}[!tb]
\centering
\caption
  {
  Mean accuracy of transfer learning on the Caltech-UCSD bird dataset (bounding box annotation provided).
  Steps represents the number of training stages.
  }
\vspace{1ex}
\scalebox{1.0}
  {
  \begin{tabular}{lcc}
  \hline
  \vspace{1ex}
  {\small\bf Method} & {\small\bf Steps} & {\small\bf Mean Accuracy}\hspace{-1ex}  \\
  \hline
   \textbf{All species (500)}  & ~ & ~ \\
  \hline
  IN-rt & 1 & 58.0\% \\
  CUB-rt & 1 & 11.4\% \\
  BS-rt & 1 & 44.8\% \\
  ~\\
  IN-CUB-ft & 2 & 68.3\% \\
  IN-BS-ft & 2 & 70.1\% \\
  IN-BS-ft-CUB-ft & 3 & \textbf{70.8}\% \\
  ~\\
  \hline
   \textbf{Non-overlapping species (366)}  & ~ & ~ \\
  \hline
  IN-BS-ft & 2 & 67.7\% \\  
  IN-BS-ft-CUB-ft & 3 & \textbf{70.5}\% \\
  ~\\
  \hline
  \textbf{Overlap (134) + Random (232)}  & ~ & ~ \\
  \hline
  IN-BS-ft & 2 & 69.5\% \\  
  \hline
  \end{tabular}
  }
 \label{table:transfer}
\end{table}

\vspace{-1ex}
\subsubsection{Transfer Learning: Experiment II}

Next we investigate transfer learning features from non-overlapping classes between two bird datasets. 
We fine-tune the pre-trained CNN using those species from the Birdsnap dataset that do not overlap with Caltech-UCSD. 
There are $134$ species that overlap and so we only use $366$ species for this experiment.

As can be seen from the second part of the Table.~\ref{table:transfer}, the result of transfer learning on Birdsnap in this setting is slightly worse with a mean accuracy of $67.7\%$.
However, if we perform progressive feature learning by learning on the target dataset (\textbf{IN-BS-ft-CUB-ft}) we obtain a mean accuracy of $70.5\%$. 
This is only $0.3\%$ worse than if we used all of the Birdsnap data and demonstrates the effectiveness of progressive feature learning.
  
\vspace{-1ex}
\subsubsection{Transfer Learning: Experiment III}

In this experiment we show the importance of overlapping classes for learning domain-generic features. 
In order to investigate if the overlapping classes play a key role to learn domain-generic features,
we fine-tuned the ImageNet model again with $134$ overlapping species
and $232$ randomly selected unique species from the Birdsnap;
this gives us $366$ species which is the number of species available in Experiment II. 
The result shows that overlapping species are important to learn domain-generic species with a mean accuracy of $69.5\%$.



\subsection{Evaluation of Subset Specific Features}

In this set of experiments we evaluate our proposed subset feature learning method on Caltech-UCSD. 
We use the same evaluation protocol as domain-generic feature learning in the previous section,
where the $DFCNN$ is used to extract features from given ground truth bounding box location of the whole bird. 
We use the acronym \textbf{SF} to indicate subset feature learning.
Based on initial experiments we set $K=6$.

Results in Table~\ref{table:bb} show that subset feature learning provides considerable improvements.
As a baseline, the results from~\cite{zhang2014part} are shown, where the features were fine-tuned on the Caltech-UCSD dataset; this equates to \textbf{IN-CUB-ft} in our terminology.
Comparing to this baseline, both of our proposed subset feature learning methods,
\textbf{IN-BS-ft-SF(SCNN)} and \textbf{IN-BS-ft-SF($k$-means)},
provide considerable improvements with mean accuracies of $72.0\%$ and $70.4\%$ respectively.
This demonstrates the effectiveness of our proposed subset feature learning technique,
and the importance of the subset selector as the SCNN approach provides an absolute performance improvement of $1.6\%$ when compared to the much simpler $k$-means approach.

\subsection{Comparison with State-of-the-Art}

In this section we demonstrate that subset feature learning can achieve state-of-the-art performance for automatic fine-grained bird classification.
Recent work in~\cite{zhang2014part} provided state-of-the-art performance on the Caltech-UCSD dataset.
This was achieved by crafting a highly accurate parts localisation model which leveraged deep convolutional features computed on bottom-up region proposals based on the RCNN framework~\cite{girshick2013rich} .
We show that if we use a similar approach but substitute their global feature vector with the feature vector obtained from subset feature learning,
then state-of-the-art performance can be achieved.


We present our results under the same setting as~\cite{zhang2014part},
where the bird detection bounding box is unknown during test time. 
This setting is fully automatic and hence more realistic. 
Since we concentrate on feature learning we use the detection results and parts features from~\cite{zhang2014part},
and substitute their global feature vector with the one we learn from subset feature learning.

\begin{figure}
  \centering
  \includegraphics[width=1\linewidth]{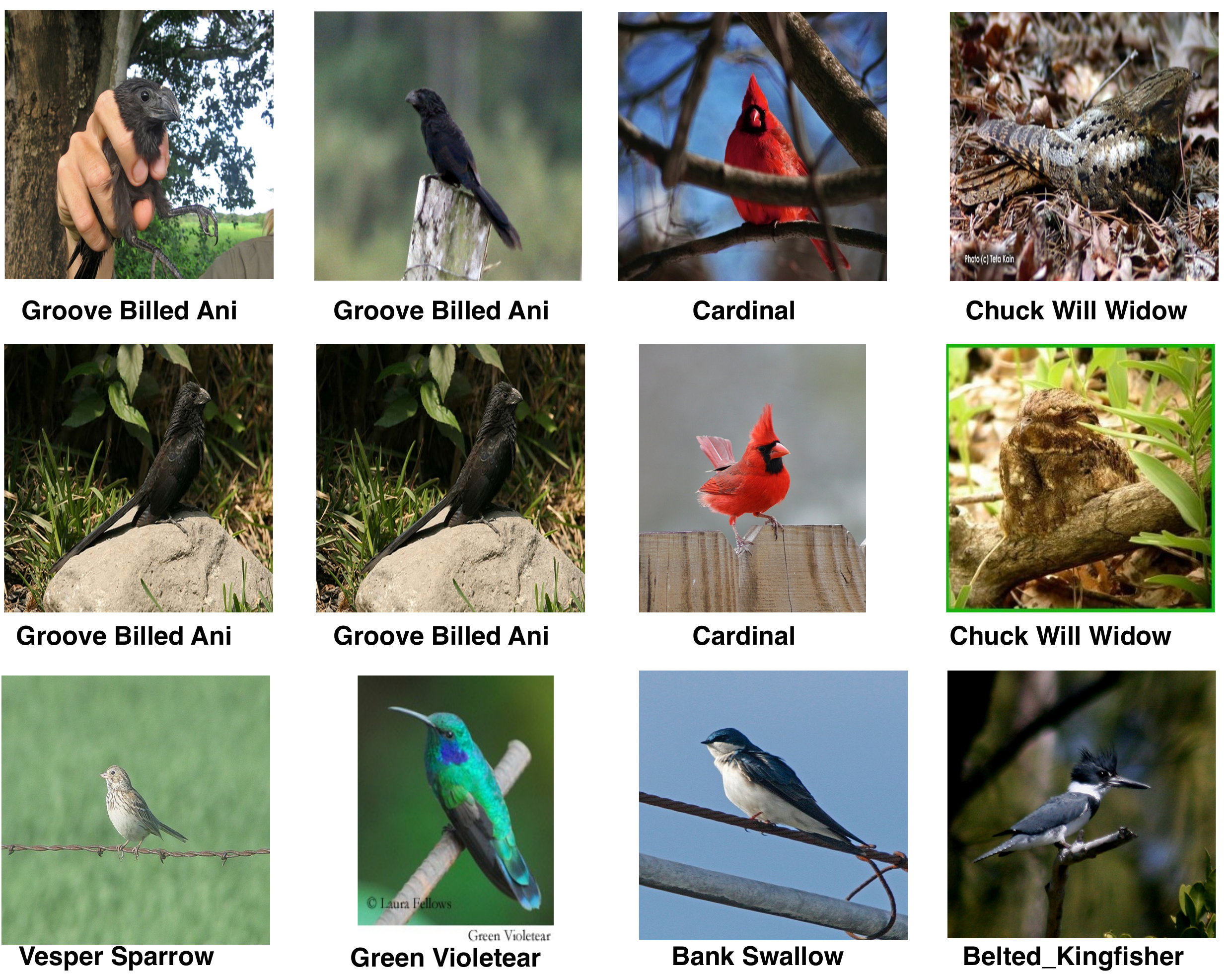}
  \caption
    {
    Qualitative comparison between our proposed method and the previous state-of-the-art approach~\cite{zhang2014part} (part-based RCNN with $\delta^{KP}$).
    The first row shows examples of test images,
    the second row shows the corresponding predicted classes from our proposed method,
    and the last row images shows the predictions using~\cite{zhang2014part}.
    It can be seen that the previous state-of-the-art approach made errors despite the large visual dissimilarities between the test image and the predicted classes.
    In contrast, the proposed approach provides the correct class labels in these cases.
    }
  \label{fig:far}
  \vspace{1ex}
  \hrule
\end{figure}

The results in Table~\ref{table:ubb} show that our proposed method achieves a mean accuracy of $77.2\%$ when we use domain-generic features and subset-specific features. 
This is a considerable improvement over the previous state-of-the-art system~\cite{zhang2014part} which achieved a mean accuracy of $73.2\%$.
An extra $0.3\%$ performance is gained when we perform progressive feature learning and fine-tune the CNN model again on the Caltech-UCSD dataset. 
Qualitative results are shown in Fig.~\ref{fig:far} which highlight instances
where the previous state-of-the-art methods provides an incorrect class label despite large visual dissimilarities.
In contrast, our approach provides the correct class label.

\begin{table}[!tb]
\centering
\caption
  {
  Mean accuracy on the Caltech-UCSD bird dataset of subset-specific features learned using subset feature learning.
  Annotated bounding boxes are used.
  }
\scalebox{1.0}
  {
  \begin{tabular}{lcc}
  \hline
  \vspace{1ex}
  {\bf Method} & {\bf Mean Accuracy}  \\
  \hline
  Fine-tuned Decaf~\cite{zhang2014part} & 68.3\% \\
  IN-BS-ft + SF(k-means) & 70.4\% \\
  IN-BS-ft + SF(SCNN) & \textbf{72.0\%} \\
  \hline
  \end{tabular}
  }
 \label{table:bb}
\end{table}

\begin{table}[!tb]
\centering
\caption
  {
  Comparison to recent results on the Caltech-UCSD bird dataset.
  Bounding boxes are not used.
  }
\scalebox{1.0}
  {
  \begin{tabular}{lcc}
  \hline
  \vspace{1ex}
  {\bf Method} & {\bf Mean Accuracy}  \\
  \hline
  DPD-DeCAF~\cite{zhang2013deformable} & 44.9\% \\
  Part-based RCNN with $\delta^{KP}$~\cite{zhang2014part} & 73.2\% \\ 
  IN-BS-ft + SF(k-means) with $\delta^{KP}$  & 76.2\% \\
  IN-BS-ft + SF(SCNN) with $\delta^{KP}$  & \textbf{77.2\%} \\
  IN-BS-ft-CUB-ft + SF with $\delta^{KP}$ & \textbf{77.5\%} \\
  \hline
  \end{tabular}
  }
 \label{table:ubb}
\end{table}

%
%

\section{Conclusion}
\label{sec:conclusion}

We have proposed a progressive transfer learning system to learn domain-generic features as well 
as subset learning to learn subset specific features.
For progressive transfer learning, we have shown that it is possible to learn domain-generic features for tasks such as fine-grained image classification.
Furthermore, we have shown that progressive transfer learning of these domain-generic features can be performed to learn target set specific features, yielding considerable improvements in accuracy.

Finally, we have presented a subset feature learning system that is able to learn subset-specific features.
Using this approach we achieve state-of-the-art performance of $77.5\%$ for fully automatic fine-grained bird image classification, the most difficult setting.
We believe our proposed method can be useful not only for fine-grained image classification, but also for improving general object recognition.
We will examine this potential in future work.



\section*{Acknowledgments}

\begin{small}
The Australian Centre for Robotic Vision is supported by the Australian Research Council via the Centre of Excellence program.
NICTA is funded by the Australian Government through the Department of Communications,
as well as the Australian Research Council through the ICT Centre of Excellence program.
\end{small}

\balance
\small
\bibliographystyle{ieee}
\bibliography{refs}

\end{document}